\documentclass{bmvc2k}
\def\arxivcopy{}
\usepackage{graphicx}
\DeclareGraphicsExtensions{.pdf,.png}
\usepackage{url}
\usepackage{ragged2e}
\usepackage{times}
\usepackage{array}
\usepackage{amsmath}
\usepackage{amssymb}
\usepackage{caption}
\usepackage{subcaption}
\usepackage[inline]{enumitem}
\usepackage{float}
\usepackage{xcolor}
\usepackage{colortbl}
\usepackage{makecell}
\usepackage{xspace}
\usepackage{etoolbox}
\usepackage{tikz}
\DeclareMathOperator*{\argmax}{arg\,max}

\newcommand{\R}{\mathbb{R}}
\newcommand{\SUM}{\emph{SH}}
\newcommand{\MAX}{\emph{MH}}

\newcommand{\OC}{\emph{1C}}
\newcommand{\TC}{\emph{10C}}
\newcommand{\RC}{\emph{RC}}
\newcommand{\TCV}{\emph{10C+rV}}
\newcommand{\RCV}{\emph{RC+rV}}
\newcommand{\RV}{\emph{rV}}

\newcommand{\order}{\emph{Order}}
\newcommand{\orderz}{\emph{Order0}}
\newcommand{\orderpp}{\emph{Order++}}
\newcommand{\VSE}{\emph{UVS}}
\newcommand{\VSEpp}{\emph{VSE++}}
\newcommand{\VSEppFt}{\emph{VSE++ (FT)}}
\newcommand{\VSEppRes}{\emph{VSE++ (ResNet)}}
\newcommand{\VSEppResFt}{\emph{VSE++ (ResNet, FT)}}
\newcommand{\VSEz}{\emph{VSE0}}
\newcommand{\VSEzFt}{\emph{VSE0 (FT)}}
\newcommand{\VSEzRes}{\emph{VSE0 (ResNet)}}
\newcommand{\VSEzResFt}{\emph{VSE0 (ResNet, FT)}}

\newcommand{\coco}{MS-COCO}
\newcommand{\fthk}{Flickr30K}

\newcounter{rowcntr}[table]
\renewcommand{\therowcntr}{\the\numexpr\thetable+1.\arabic{rowcntr}}

\newcommand{\numrow}{\refstepcounter{rowcntr}\therowcntr}

\AtBeginEnvironment{tabular}{\setcounter{rowcntr}{0}}

\newcolumntype{H}{>{\setbox0=\hbox\bgroup}c<{\egroup}@{}}

\newcommand{\comment}[1]{}
\makeatletter
\def\blfootnote{\gdef\@thefnmark{}\@footnotetext}
\makeatother

\title{VSE++: Improving Visual-Semantic
Embeddings with Hard Negatives}

\addauthor{Fartash Faghri}{faghri@cs.toronto.edu}{1}
\addauthor{David J. Fleet}{fleet@cs.toronto.edu}{1}
\addauthor{Jamie Ryan Kiros$^*$}{kiros@google.com}{2}
\addauthor{Sanja Fidler}{fidler@cs.toronto.edu}{1}

\addinstitution{%
Department of Computer Science,\\
University of Toronto \\
and\\
Vector Institute
}
\addinstitution{%
    Google Brain Toronto
}

\runninghead{Faghri \etal}{VSE++: Improving Visual-Semantic Embeddings with H.  
N.}

\def\etal{\emph{et al}\bmvaOneDot}

\begin{document}

\maketitle

\vspace{-1mm}
\begin{abstract}
    We present a new technique for learning visual-semantic embeddings for 
    cross-modal retrieval.  Inspired by hard negative mining, the use of hard 
    negatives in structured prediction, and ranking loss functions, we 
    introduce a simple change to common loss functions used for multi-modal 
    embeddings.  That, combined with fine-tuning and use of augmented data, 
    yields significant gains in retrieval performance.  We showcase our 
    approach, \VSEpp, on \coco{} and \fthk{} datasets, using ablation studies 
    and comparisons with existing methods.  On \coco{} our approach outperforms 
    state-of-the-art methods by $8.8\%$ in caption retrieval and $11.3\%$ in 
    image retrieval (at R@$1$).\blfootnote{$^*$Work done while a Ph.D. student 
    at the University of Toronto.}
\end{abstract}

\section{Introduction}

Joint embeddings enable a wide range of tasks in image, video and language 
understanding. Examples include shape-image embeddings (\cite{li2015joint}) 
for shape inference, bilingual word embeddings (\cite{zou2013bilingual}), 
human pose-image embeddings for 3D pose inference (\cite{li2015maximum}), 
fine-grained recognition (\cite{reed2016learning}), zero-shot learning 
(\cite{frome2013devise}), and modality conversion via synthesis 
(\cite{reed2016generative,reed2016learning}).  Such embeddings entail 
mappings from two (or more) domains into a common vector space in 
which semantically associated inputs (e.g., text and images) are mapped to 
similar locations.  The embedding space thus represents the underlying 
domain structure, where location and often direction are 
semantically meaningful.

{\em Visual-semantic embeddings} have been central to
image-caption retrieval and 
generation~\cite{kiros2014unifying,karpathy2015deep}, and visual 
question-answering~\cite{Malinowski15}.  One approach to visual 
question-answering, for example, is to first describe an image by a set of 
captions, and then to find the nearest caption in response to a question 
(\cite{agrawal2017vqa,zitnick2016measuring}). For image synthesis 
from text, one could map from text to the joint embedding space, and then
back to image space (\cite{reed2016generative,reed2016learning}).

Here we focus on visual-semantic embeddings for cross-modal retrieval; 
i.e.\ the retrieval of images given captions, 
or of captions for a query image.  As is common in retrieval, 
we measure performance by R@$K$, i.e., recall at $K$ -- the fraction of 
queries for which the correct item is retrieved in the closest $K$ points 
to the query in the embedding space ($K$ is usually a small integer, 
often $1$).  More generally, retrieval is a natural way to assess the 
quality of joint embeddings for image and language data (\cite{hodosh2013framing}).

The basic problem is one of ranking; the correct target(s) 
should be closer to the query than other items in the corpus, not unlike
{\em learning to rank}\/ problems (e.g., \cite{li2014learning}), and 
max-margin structured prediction \cite{chapelle2007large,le2007direct}.
The formulation and model architecture in this paper are most closely 
related to those of \cite{kiros2014unifying}, learned with a triplet ranking 
loss.  In contrast to that work, we advocate a novel loss, the use of 
augmented data, and fine-tuning, which, together, produce a significant 
increase in caption retrieval performance over the baseline ranking loss on 
well-known benchmark data.  We outperform the best reported result on 
\coco{} by almost $9\%$.  We also show that the benefit of a more 
powerful image encoder, with fine-tuning, is amplified with 
the use of our stronger loss function.
% To ensure reproducibility, our code will be made publicly available.
We refer to our model as \VSEpp{}.
To ensure reproducibility, our code is publicly 
available~\footnote{\url{https://github.com/fartashf/vsepp}}.

Our main contribution is to incorporate hard negatives in the loss 
function.  This was inspired by the use of hard negative mining in 
classification tasks (\cite{dalal2005histograms, 
felzenszwalb2010object, malisiewicz2011ensemble}), and by the use of
hard negatives for improving image embeddings for face recognition 
(\cite{schroff2015facenet, wu2017sampling}). Minimizing a loss function using 
hard negative mining is equivalent to minimizing a modified non-transparent 
loss function with uniform sampling.  We extend the idea with the explicit 
introduction of hard negatives in the loss for multi-modal embeddings, 
without any additional cost of mining.

We also note that our formulation complements other recent articles that 
propose new architectures or similarity functions for this problem. To this 
end, we demonstrate improvements to \cite{vendrov2015order}.
Among other methods that could be improved with a modified loss, 
\cite{wang2017learning} propose an embedding network to fully 
replace the similarity function used for the ranking loss.  An attention 
mechanism on both images and captions is used by \cite{nam2016dual}, where the 
authors sequentially and selectively focus on a subset of words and image 
regions to compute the similarity.  In \cite{huang2016instance}, the authors 
use a multi-modal context-modulated attention mechanism to compute the 
similarity between images and captions. Our proposed loss function and 
triplet sampling could be extended and applied to other such problems.

\vspace{-3mm}
\section{Learning Visual-Semantic Embeddings}

%\subsection{Image-Caption Retrieval}

For image-caption retrieval the query is a caption and the task is to 
retrieve the most relevant image(s) from a database. Alternatively, the 
query may 
be an image, and the task is to retrieves relevant captions.  The goal is to 
maximize recall at $K$ (R@$K$), i.e., the fraction of queries for which the 
most relevant item is ranked among the top $K$ items returned.

Let $S=\{(i_n, c_n)\}^N_{n=1}$ be a training set of image-caption pairs.  
We refer to $(i_n, c_n)$ as {\em positive pairs}\/ and $(i_n, c_{m\neq n})$ as 
{\em negative pairs}\/; i.e., the most relevant caption to the image $i_n$ is 
$c_n$ and for caption $c_n$, it is the image $i_n$.  We define a similarity 
function $s(i, c)\in \R$ that should, ideally, give higher similarity 
scores to positive pairs than negatives.  In caption retrieval, the query 
is an image and we rank a database of captions based on the similarity 
function; i.e., R@$K$ is the percentage of queries for which the positive 
caption is ranked among the top $K$ captions using $s(i, c)$.  Likewise 
for image retrieval.  In what follows the similarity function is defined 
on the joint embedding space.  This differs from other formulations, such 
as \cite{wang2017learning}, which use a similarity network to directly 
classify an image-caption pair as matching or non-matching.

\vspace{-3mm}
\subsection{Visual-Semantic Embedding}

Let $\phi(i; \theta_\phi)\in \R^{D_\phi}$ be a feature-based representation 
computed from image $i\,$ (e.g.\ the representation before logits in 
VGG19~(\cite{simonyan2014very}) or ResNet152~(\cite{he2016deep})).  
Similarly, let $\psi(c; \theta_\psi) \in \R^{D_\psi}$ be a representation
of caption $c$ in a caption embedding space (e.g.\ a GRU-based text 
encoder).  Here, $\theta_\phi$ and $\theta_\psi$ denote model 
parameters for the respective mappings to these initial 
image and caption representations.

Then, let the mappings into the {\em joint embedding space}\/ 
be defined by linear projections:
\begin{eqnarray}
    f(i; W_f, \theta_\phi) & = & W_f^T \phi(i; \theta_\phi)\\
    g(c; W_g, \theta_\psi) & = & W_g^T \psi(c; \theta_\psi) % \\
\end{eqnarray}
where $W_f\in\R^{D_\phi\times D}$ and $W_g\in\R^{D_\psi\times D}$.
We further normalize $f(i; W_f, \theta_\phi)$, and $g(c; W_g, \theta_\psi)$, 
to lie on the unit hypersphere.  Finally, we define the similarity 
function in the joint embedding space to be the usual inner product:
\begin{align}
s(i,c) = f(i; W_f, \theta_\phi)\cdot g(c; W_g, \theta_\psi)\,.
\end{align}
Let $\theta=\{W_f,W_g,\theta_\psi\}$ be the model parameters.  If we also
fine-tune the image encoder, then we would also include $\theta_\phi$ in 
$\theta$.

Training entails the minimization of empirical loss with respect to $\theta$, 
i.e., the cumulative loss over training data ${S=\{(i_n, c_n)\}^N_{n=1}}$:
\begin{eqnarray}
e(\theta, S)=\tfrac{1}{N}\sum^N_{n=1} \ell(i_n, c_n)
\end{eqnarray}
where $\ell(i_n, c_n)$ is a suitable loss function for a single training 
exemplar.
Inspired by the use of a triplet loss for image retrieval
(e.g., \cite{frome2007learning,chechik2010large}), recent approaches
to joint visual-semantic embeddings have used a hinge-based triplet ranking 
loss \cite{kiros2014unifying, karpathy2015deep, ZhuICCV15, 
socher2014grounded}:
\begin{equation}
    \ell_{SH}(i, c) ~=~
    \sum_{\hat{c}} [\alpha - s(i,c) + s(i,\hat{c})]_+ 
 \, +\, \sum_{\hat{i}} [\alpha - s(i,c) + s(\hat{i},c)]_+\,,
    \label{eq:contrastive}
\end{equation}
where $\alpha$ serves as a margin parameter, and $[x]_+ \equiv \max(x, 0)$.  
This hinge loss comprises two symmetric terms.  The first sum is taken over all 
negative captions $\hat{c}$, given query $i$.  The second is taken over all 
negative images $\hat{i}$, given caption $c$.  Each term is proportional to the 
expected loss (or {\em violation}\/) over sets of negative samples.  If $i$ and 
$c$ are closer to one another in the joint embedding space than to any 
negative, by the margin $\alpha$, the hinge loss is zero.  In practice, for 
computational efficiency, rather than summing over all  negatives in the 
training set, it is common to only sum over (or randomly sample) the negatives 
in a mini-batch of stochastic gradient descent \cite{kiros2014unifying, 
socher2014grounded, karpathy2015deep}.  The runtime complexity of computing 
this loss approximation is quadratic in the number of image-caption pairs in 
a mini-batch.

Of course there are other loss functions that one might consider.
One  is a pairwise hinge loss in which elements of positive pairs
are encouraged to lie within a hypersphere of radius $\rho_1$ in 
the joint embedding space,
while negative pairs should be no closer than $\rho_2 > \rho_1$.
This is problematic as it constrains the structure of the latent space more 
than does the ranking loss, and it entails the use of two hyper-parameters 
which can be very difficult to set.
Another possible approach is to use Canonical Correlation Analysis to learn 
$W_f$ and $W_g$, thereby trying to preserve correlation between the text and 
images in the joint embedding (e.g., 
\cite{klein2015associating,eisenschtat2016linking}).  By comparison, when 
measuring performance as R@$K$, for small $K$, a correlation-based loss will 
not give sufficient influence to the embedding of negative items in the local 
vicinity of positive pairs, which is critical for R@$K$.  

\subsection{Emphasis on Hard Negatives}
\label{sec:hard_neg}

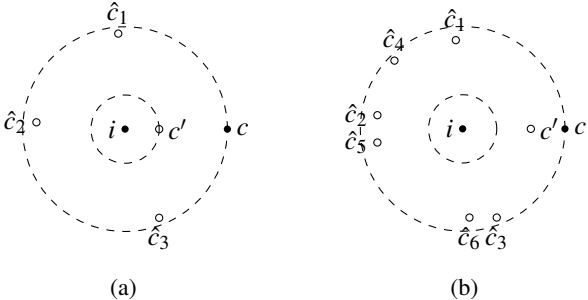
\begin{figure*}[t!]

%\vspace*{-.5cm}
    \hspace*{-1cm}
    \centering
    \def\xx{5}
    \def\rr{1.5}
    \def\nn{.5}
    \begin{subfigure}[b]{.18\textwidth}
        \hspace*{-.6cm}
        \begin{tikzpicture}[scale=0.9]
            \draw[dashed] (0, 0) circle (\rr);
            \fill (0,0) circle (1.5pt) node[left] {$i$};
            \fill (\rr,0) circle (1.5pt) node[right] {$c$};
            \draw[dashed] (0, 0) circle (.5);
            \draw (+\nn,0) circle (1.5pt) node[right] {$c^\prime$};

            % \draw (.3,\nn+.3) circle (1.5pt) node[above] {$\hat{c}_1$};
            % \draw (-\nn-.4,-.2) circle (1.5pt) node[left] {$\hat{c}_2$};
            % \draw (-.6,-\nn) circle (1.5pt) node[below] {$\hat{c}_3$};

            \draw (-.1,\rr-.1) circle (1.5pt) node[above] {$\hat{c}_1$};
            \draw (+-\rr+.2,+.1) circle (1.5pt) node[left] {$\hat{c}_2$};
            \draw (0.5,-\rr+.2) circle (1.5pt) node[below] {$\hat{c}_3$};
        \end{tikzpicture}
        \caption{}\label{fig:hardneg}
    \end{subfigure}
    %\qquad
    \hspace*{2cm}
    \begin{subfigure}[b]{.18\textwidth}
        \hspace*{-.6cm}
        \begin{tikzpicture}[scale=0.9]
            \draw[dashed] (0, 0) circle (\rr);
            \fill (0,0) circle (1.5pt) node[left] {$i$};
            \fill (\rr,0) circle (1.5pt) node[right] {$c$};
            \draw[dashed] (0, 0) circle (.5);
            \draw (\rr-.5,0) circle (1.5pt) node[right] {$c^\prime$};

            \draw (-.1,\rr-.1-.1) circle (1.5pt) node[above] {$\hat{c}_1$};
            \draw (-\rr+.25,+.2) circle (1.5pt) node[left] {$\hat{c}_2$};
            \draw (0.5,-\rr+.2) circle (1.5pt) node[below] {$\hat{c}_3$};
            \draw (-1.,\rr-.5) circle (1.5pt) node[above] {$\hat{c}_4$};
            \draw (-\rr+.25,-.2) circle (1.5pt) node[left] {$\hat{c}_5$};
            \draw (0.1,-\rr+.2) circle (1.5pt) node[below] {$\hat{c}_6$};
        \end{tikzpicture}
        \captionsetup{margin=.5cm}
        \caption{}\label{fig:softneg}
    \end{subfigure}
    \caption{\small An illustration of typical positive pairs and the nearest negative 
    samples. Here assume similarity score is the negative distance. Filled 
    circles show a positive pair $(i,c)$, while empty circles are negative 
    samples for the query $i$.  The dashed circles on the two sides are drawn 
    at the same radii.  Notice that the hardest negative sample $c^\prime$ is 
    closer to $i$ in \subref{fig:hardneg}. Assuming a zero margin, 
    \subref{fig:softneg} has a higher loss with the \SUM{} loss compared to 
    \subref{fig:hardneg}.  The \MAX{} loss assigns a higher loss to 
    \subref{fig:hardneg}.}
    \label{fig:examples}
\end{figure*}

%%% referencing structured prediction
Inspired by common loss functions used in structured prediction 
(\cite{tsochantaridis2005large, yu2009learning, felzenszwalb2010object}), we 
focus on hard negatives for training, i.e., the negatives closest to each 
training query.  This is particularly relevant for retrieval since it is the 
hardest negative that determines success or failure as measured by R@$1$.  

% Given a pair $(i, c)$, we denote the hardest negative samples in the training 
% set by $i^\prime$ and $c^\prime$ as follows:
% \begin{eqnarray}
%     i^\prime&=&\argmax_{j\neq i} s(j,c)\\
%     c^\prime&=&\argmax_{d\neq c} s(i,d)\,,
% \end{eqnarray}
% i.e.,  $i^\prime$ is the most similar negative image sample to $c$, and
% $c^\prime$ is the most similar negative caption sample to $i$.  

Given a positive pair $(i, c)$, the hardest negatives are given by 
$i^\prime=\argmax_{j\neq i} s(j,c)$ and $c^\prime=\argmax_{d\neq c} s(i,d)$.
% (i.e. $c^\prime$ is the most similar negative caption sample to $i$).  
% (i.e.  $i^\prime$ is the most similar negative image sample to $c$), and
To emphasize hard negatives we define our loss as
\begin{equation}
    \ell_{MH}(i, c)
    ~=~ \max_{c^\prime} \left[\alpha + s(i,c^\prime) - s(i,c)\right]_+
    \, +\, \max_{i^\prime} \left[\alpha + s(i^\prime,c) -s(i,c)\right]_+ ~ .
    \label{eq:rank_loss}
\end{equation}
Like Eq.~\ref{eq:contrastive}, this loss comprises two terms, 
one with $i$ and one with $c$ as queries.  Unlike Eq.~\ref{eq:contrastive}, 
this loss is specified in terms of the hardest negatives,  $c^\prime$ and 
$i^\prime$. We refer to the loss in Eq.~\ref{eq:rank_loss} 
as {\em Max of Hinges (\MAX{})}\/ loss, and the loss in 
Eq.~\ref{eq:contrastive} as {\em Sum of Hinges (\SUM{})}\/ loss. There is 
a spectrum of loss functions from the \SUM{} loss to the \MAX{} loss. In the 
\MAX{} loss, the winner takes all the gradients, where instead we use 
re-weighted gradients of all the triplets. We only discuss the \MAX{} loss as 
it was empirically found to perform the best.

One case in which the \MAX{} loss is superior to \SUM{} is when multiple 
negatives with small violations combine to dominate the \SUM{} loss.  
For example, Fig.~\ref{fig:examples} depicts a positive pair
together with two sets of negatives.  In Fig.~\ref{fig:hardneg}, a 
single negative is too close to the query, which may require
a significant change to the mapping.  
However, any training step that pushes the hard negative 
away, might cause a number of small violating negatives, as 
in Fig.~\ref{fig:softneg}. Using the \SUM{} loss, these `new' negatives 
may dominate the loss, so the model is pushed back to the first 
example in Fig.~\ref{fig:hardneg}.  This may create local minima in the 
\SUM{} loss that may not be as problematic for the \MAX{} loss, which focuses
on the hardest negative.

For computational efficiency, instead of finding the hardest negatives in the 
entire training set, we find them within each mini-batch.  
This has the same quadratic 
complexity as the complexity of the \SUM{} loss.  With random 
sampling of the mini-batches, this approximation yields other advantages.  One is 
that there is a high probability of getting hard negatives that are harder than 
at least $90\%$ of the entire training set.  Moreover, the loss is potentially 
robust to label errors in the training data because the probability of sampling 
the hardest negative over the entire training set is somewhat low.

\subsection{Probability of Sampling the Hardest Negative}
\label{sec:prob_hard_neg}

Let $S=\{(i_n, c_n)\}^N_{n=1}$ denote a training set of image-caption pairs,
and let $C=\{c_n\}$ denote the set of captions.  Suppose we draw $M$ samples 
in a mini-batch, $Q=\{(i_m, c_m)\}^M_{m=1}$, from $S$.  Let the permutation, 
$\pi_m$, on $C$ refer to the rankings of captions according to the similarity 
function $s(i_m,c_n)$ for $c_n\in S\setminus\{c_m\}$. We can assume 
permutations, $\pi_m$, are uncorrelated.

Given a query image, $i_m$, we are interested in the probability of getting 
no captions from the $90$th percentile of $\pi_m$ in the mini-batch. 
Assuming IID samples, this probability is simply $.9^{(M-1)}$, the probability 
that no sample in the mini-batch is from the $90$th percentile. 
This probability tends to zero exponentially fast, falling below $1\%$ 
for $M\geq 44$. Hence, for large enough mini-batchs, with high probability 
we sample negative captions that are harder 
than $90\%$ of the entire training set.
The probability for the $99.9$th percentile of $\pi_m$ tends to zero 
more slowly; it falls below $1\%$ for $M\geq 6905$, which is a relatively 
large mini-batch. 

While we get strong signals  by randomly 
sampling negatives within mini-batches, such sampling also provides 
some robustness to outliers, such as negative captions that 
better describe an image compared to the ground-truth caption.
Mini-batches as small as $128$ can provide strong 
enough training signal and robustness to label errors. Of course by 
increasing the mini-batch size, we get harder negative examples and possibly 
a stronger training signal.  However, by increasing the mini-batch size, we lose 
the benefit of SGD in finding good optima and exploiting the gradient noise.  
This can lead to getting stuck in local optima or as observed by 
\cite{schroff2015facenet}, extremely long training time.

\section{Experiments}

\begin{table*}[t!]
 %\begin{table*}[t]
\resizebox{\linewidth}{!}{
    \centering
     \begin{tabular}{c|c|c|HccccH|ccccH}
         \# &
         {\bf Model} & {\bf Trainset} & {\bf R Sum}
         & \multicolumn{5}{c|}{Caption Retrieval}
         & \multicolumn{5}{c}{Image Retrieval}\\
         & & & &
         {\bf R@1} & {\bf R@5} & {\bf R@10} & {\bf Med r} & {\bf Mean r} &
         {\bf R@1} & {\bf R@5} & {\bf R@10} & {\bf Med r} & {\bf Mean r}\\
         \hline
         & & \multicolumn{12}{c}{{\cellcolor[gray]{0.8}\bf 1K Test Images}}\\
         \hline
         % \VSE{}~\cite{kiros2014unifying} & \OC{} (1 fold) &
         % $344.1$ &
         % $33.8$ & $67.7$ & $82.1$ & $3$ & - &
         % $25.9$ & $60.0$ & $74.6$ & $4$ & - \\
         \numrow{}\label{coco:VSE}&
         \VSE{}~(\cite{kiros2014unifying}, GitHub) & \OC{} (1 fold) &
         $382.5$ &
         $43.4$ & $75.7$ & $85.8$ & $2$ & - &
         $31.0$ & $66.7$ & $79.9$ & $3$ & - \\
         \numrow{} &
         \order{}~(\cite{vendrov2015order})& \TCV{} &
         - &
         $46.7$ & - & $88.9$ & $2.0$ & $5.7$ &
         $37.9$ & - & $85.9$ & $2.0$ & $8.1$\\
         \numrow{} &
         Embedding Net~(\cite{wang2017learning}) & \TCV{} &
         - &
         $50.4$ & $79.3$ & $69.4$ & - & - &
         $39.8$ & $75.3$ & $86.6$ & - & -
         \\
         \numrow{}\label{coco:smLSTM} &
         sm-LSTM~(\cite{huang2016instance}) &? &
         $431.8$ &
         $53.2$ & $83.1$ & $91.5$ & $\mathbf{1}$ & - &
         $40.7$ & $75.8$ & $87.4$ & $2$ & -
         \\
         \numrow{}\label{coco:twoway} &
         2WayNet~(\cite{eisenschtat2016linking}) & \TCV{} &
         - &
         $55.8$ & $75.2$ & - & - & - &
         $39.7$ & $63.3$ & - & - & -
         \\
         \hline
         % coco_vse++
         % --data_name coco_precomp --max_violation
         \numrow{}\label{coco:VSEppOC} &
         \VSEpp{} & \OC{} (1 fold) &
         $386.5$ &
         $43.6$ & $74.8$ & $84.6$ & $2.0$ & $8.8$ &
         $33.7$ & $68.8$ & $81.0$ & $3.0$ & $12.9$\\

         % coco_vse++_vggfull
         % --data_name coco --cnn_type vgg19 --max_violation
         \numrow{}\label{coco:VSEppRC} &
         \VSEpp{} & \RC &
         $410.3$ &
         $49.0$ & $79.8$ & $88.4$ & $1.8$ & $6.5$ &
         $37.1$ & $72.2$ & $83.8$ & $2.0$ & $10.8$\\

         % coco_vse++_vggfull_restval
         % --data_name coco --cnn_type vgg19 --max_violation --use_restval
         \numrow{}\label{coco:VSEppRCV} &
         \VSEpp{} & \RCV{} &
         $423.1$ &
         $51.9$ & $81.5$ & $90.4$ & $\mathbf{1.0}$ & $5.8$ &
         $39.5$ & $74.1$ & $85.6$ & $2.0$ & $10.0$\\

         % coco_vse++_vggfull_restval_finetune
         % --data_name coco --cnn_type vgg19 --max_violation --use_restval 
         %  --finetune --num_epochs 15 --learning_rate .00002 --resume
         \numrow{}\label{coco:VSEppFt} &
         \VSEppFt{} & \RCV{} &
         $450.9$ &
         $57.2$ & $86.0$ & $93.3$ & $\mathbf{1.0}$ & $4.2$ &
         $45.9$ & $79.4$ & $89.1$ & $2.0$ & $8.5$\\

         % coco_vse++_resnet_restval
         % --data_name coco --cnn_type resnet152 --use_restval --max_violation
         \numrow{}\label{coco:VSEppRes} &
         \VSEppRes{} & \RCV{} &
         $446.8$ &
         $58.3$ & $86.1$ & $93.3$ & $\mathbf{1.0}$ & $4.2$ &
         $43.6$ & $77.6$ & $87.8$ & $2.0$ & $7.8$
         \\

         % coco_vse++_resnet_restval_finetune
         % --data_name coco --cnn_type resnet152 --use_restval --max_violation 
         %  --finetune --num_epochs 15 --learning_rate .00002 --resume
         \numrow{}\label{coco:VSEppResFt} &
         \VSEppResFt{} & \RCV{} &
         $\mathbf{478.6}$ &
         $\mathbf{64.6}$ & $\mathbf{90.0}$ & $\mathbf{95.7}$ & $\mathbf{1.0}$ 
         & $\mathbf{3.4}$ &
         $\mathbf{52.0}$ & $\mathbf{84.3}$ & $\mathbf{92.0}$ & $\mathbf{1.0}$ 
         & $\mathbf{6.1}$
         \\

        \hline
         & & \multicolumn{12}{c}{{\cellcolor[gray]{0.8}\bf 5K Test Images}}\\
         \hline
         \numrow{} &
         \order{}~(\cite{vendrov2015order})& \TCV{} &
         - &
         $23.3$ & - & $65.0$ & $5.0$ & $24.4$ &
         $18.0$ & - & $57.6$ & $7.0$ & $35.9$\\

         % coco_vse++_vggfull_restval_finetune
         \numrow{}\label{coco:VSEppFt5} &
         \VSEppFt{} & \RCV{} &
         $312.4$ &
         $32.9$ & $61.7$ & $74.7$ & $3.0$ & $16.9$ &
         $24.1$ & $52.8$ & $66.2$ & $5.0$ & $38.1$\\

         % coco_vse++_resnet_restval_finetune
         \numrow{}\label{coco:VSEppResFt5} &
         \VSEppResFt{} & \RCV{} &
         $\mathbf{355.8}$ &
         $\mathbf{41.3}$ & $\mathbf{71.1}$ & $\mathbf{81.2}$ & $\mathbf{2.0}$ 
         & $\mathbf{12.4}$ &
         $\mathbf{30.3}$ & $\mathbf{59.4}$ & $\mathbf{72.4}$ & $\mathbf{4.0}$ 
         & $\mathbf{26.0}$
        \\[-2mm]
    \end{tabular}
    }
    \vspace{.2cm}
    \caption{Results of experiments on \coco{}.}
    \label{tb:coco}
    \vspace{-.4cm}
%\end{table*}

\end{table*}
%\bigskip

Below we perform experiments with our approach, \VSEpp{}, comparing it to 
a baseline formulation with $\SUM{}$ loss, denoted \VSEz{}, and other 
state-of-the-art approaches. Essentially, the baseline formulation, \VSEz{}, 
is similar to that in \cite{kiros2014unifying}, denoted  \VSE{}.  

% \david{The issue here is not that we can complement other formulations by 
% that some of the formulations against which we compare are much more 
% complicated?  Perhaps we should move the next two sentences to where we 
% contrast our results with other formulations, or to the section on order 
% embeddings.}

We experiment with two image encoders: VGG19 by~\cite{simonyan2014very} and 
ResNet152 by~\cite{he2016deep}.  In what follows, we use VGG19 unless specified 
otherwise.  As in previous work we extract image features directly from FC$7$, 
the penultimate fully connected layer.
% We thus explicitly identify variations where we use ResNet152 instead.  
The dimensionality of the image embedding, $D_\phi$, 
is $4096$ for VGG19 and $2048$ for ResNet152.

In more detail, we first resize the image to $256\times256$, and then 
use either a single crop of size $224\times224$ or the mean of feature vectors 
for multiple crops of similar size. We refer to training with one center crop 
as \OC{}, and training with $10$ crops at fixed locations as \TC{}. These image 
features can be pre-computed once and reused. We also experiment with using 
a single random crop, denoted by \RC{}.  For \RC{}, image features are computed 
on the fly. Recent works have mostly used \RC{}/\TC{}. In our preliminary 
experiments, we did not observe significant differences between \RC{}/\TC{}. As 
such, we perform  most experiments with \RC{}.

For the caption encoder, we use a GRU similar to the one used in 
\cite{kiros2014unifying}.  We set the dimensionality of the GRU, $D_\psi$, and 
the joint embedding space, $D$, to $1024$. The dimensionality of the word 
embeddings that are input to the GRU is set to $300$.

We further note that in \cite{kiros2014unifying}, the caption embedding is 
normalized, while the image embedding is not. Normalization of both vectors 
means that the similarity function is cosine similarity.  In \VSEpp{} we 
normalize both vectors. Not normalizing the image embedding changes the 
importance of samples. In our experiments, not normalizing the image 
embedding  helped the baseline, \VSEz{}, to find a better solution. 
However, \VSEpp{} is not significantly affected by this normalization.

\subsection{Datasets}

We evaluate our method on the Microsoft COCO dataset (\cite{lin2014microsoft}) 
and the \fthk{} dataset (\cite{young2014image}). \fthk{} has a standard 
$30,000$ images for training. Following \cite{karpathy2015deep}, we use $1000$ 
images for validation and $1000$  images for testing. We also use the splits of 
\cite{karpathy2015deep} for \coco{}. In this split, the training set contains 
$82,783$ images, $5000$ validation and $5000$ test images.  However, there are 
also $30,504$ images that were originally in the validation set of \coco{} but 
have been left out in this split. We refer to this set as \RV{}. Some papers 
use \RV{} for training ($113,287$ training images in total) to further improve 
accuracy.  We report results using both training sets.  Each image comes with 
$5$ captions.  The results are reported by either averaging over $5$ folds of 
$1K$ test images or testing on the full $5K$ test images.

%Recent works have mostly used \RCV{}/\TCV{} on \coco{}, and \RC{}/\TC{} on 
%other datasets.  We did not observe significant improvements from \TC{} to 
%\RC{}, so we used \RC{} in our experiments. Unfortunately, not all previous 
%work provide such implementation details.  \cite{huang2016instance} and 
%\cite{nam2016dual} use features from a collection of candidates or regions.  
%They have not specified whether they have used precomputed features 
%(equivalent to \TC{}) or it is computed on the fly (equivalent to \RC).  Also 
%\cite{huang2016instance} does not say whether they used \RV{}.

\subsection{Details of Training}
\label{sec:train_detail}
We use the Adam optimizer~\cite{kingma2014adam}.  Models are trained 
for at most $30$ epochs.  Except for fine-tuned models,  we start 
training with learning rate $0.0002$ for $15$ epochs, and then lower the 
learning rate to $0.00002$ for another $15$ epochs.  The fine-tuned models are 
trained by taking a model trained for $30$ epochs with a fixed image 
encoder, and then training it for $15$ epochs with a learning rate of $0.00002$.  
We set the margin to $0.2$ for most experiments.  We use a mini-batch 
size of $128$ in all  experiments.  Notice that since the size of the 
training set for different models is different, the actual number of 
iterations in each epoch can vary. For evaluation on the test set, 
we tackle over-fitting by choosing the snapshot of the model that 
performs best on the validation set.  The best snapshot is selected based 
on the sum of the recalls on the validation set.

% \david{Would it be better to say that we training until performance on 
% a validation set stops improving.  Ie don't say we trained for 
% 30 epochs above, but rather we trained for at most 30, untill testing on 
% validation set showed decreased performance and hence signs of overfitting.}

% We did not see a difference in the performance by changing the mini-batch 
% size if we run for the full $30$ epochs.  

\subsection{Results on \coco{}}

\begin{table*}[t!]
 % \begin{table*}[t]
 \resizebox{\linewidth}{!}{
\centering
     \begin{tabular}{c|c|c|HccccH|ccccH}
         \# &
         {\bf Model} & {\bf Trainset} & {\bf R Sum}
         & \multicolumn{5}{c|}{Caption Retrieval}
         & \multicolumn{5}{c}{Image Retrieval}\\
         & & & &
         {\bf R@1} & {\bf R@5} & {\bf R@10} & {\bf Med r} & {\bf Mean r} &
         {\bf R@1} & {\bf R@5} & {\bf R@10} & {\bf Med r} & {\bf Mean r}\\
         \hline
         % coco_vse0
         % --data_name coco_precomp --no_imgnorm
         \numrow{}\label{coco:VSEzOC} &
         \VSEz{} & \OC{} (1 fold) &
         $383.2$ &
         $43.2$ & $73.9$ & $85.0$ & $2.0$ & $7.6$ &
         $33.0$ & $67.4$ & $80.7$ & $3.0$ & $11.3$\\

         {}\ref{coco:VSEppOC} &
         \VSEpp{} & \OC{} (1 fold) &
         $386.5$ &
         $43.6$ & $74.8$ & $84.6$ & $2.0$ & $8.8$ &
         $33.7$ & $68.8$ & $81.0$ & $3.0$ & $12.9$\\

         % % coco_vse++_vggfull
         % % --data_name coco --cnn_type vgg19 --max_violation
         % \VSEpp{} & \RC &
         % $410.3$ &
         % $49.0$ & $79.8$ & $88.4$ & $1.8$ & $6.5$ &
         % $37.1$ & $72.2$ & $83.8$ & $2.0$ & $10.8$\\

         % coco_vse0_vggfull
         % --data_name coco --cnn_type vgg19 --no_imgnorm
         \numrow{}\label{coco:VSEzRC} &
         \VSEz{} & \RC &
         $390.0$ &
         $43.1$ & $77.0$ & $87.1$ & $2.0$ & $6.5$ &
         $32.5$ & $68.3$ & $82.1$ & $3.0$ & $9.5$\\

         {}\ref{coco:VSEppRC} &
         \VSEpp{} & \RC &
         $410.3$ &
         $49.0$ & $79.8$ & $88.4$ & $1.8$ & $6.5$ &
         $37.1$ & $72.2$ & $83.8$ & $2.0$ & $10.8$\\

         % coco_vse0_vggfull_restval
         % --data_name coco --cnn_type vgg19 --use_restval --no_imgnorm
         \numrow{}\label{coco:VSEzRCV} &
         \VSEz{} & \RCV{} &
$402.7$ &
$46.8$ & $78.8$ & $89.0$ & $1.8$ & $6.1$ &
$34.2$ & $70.4$ & $83.6$ & $2.6$ & $8.6$
         \\
         {}\ref{coco:VSEppRCV} &
         \VSEpp{} & \RCV{} &
         $423.1$ &
         $51.9$ & $81.5$ & $90.4$ & $\mathbf{1.0}$ & $5.8$ &
         $39.5$ & $74.1$ & $85.6$ & $2.0$ & $10.0$\\

         % coco_vse0_vggfull_restval_finetune
         % --data_name coco --cnn_type vgg19 --use_restval --no_imgnorm 
         %  --finetune --num_epochs 15 --learning_rate .00002 --resume
         \numrow{}\label{coco:VSEzFt} &
         \VSEzFt{} & \RCV{} &
$424.3$ &
$50.1$ & $81.5$ & $90.5$ & $1.6$ & $5.5$ &
$39.7$ & $75.4$ & $87.2$ & $2.0$ & $7.1$
         \\
         {}\ref{coco:VSEppFt} &
         \VSEppFt{} & \RCV{} &
         $450.9$ &
         $57.2$ & $86.0$ & $93.3$ & $\mathbf{1.0}$ & $4.2$ &
         $45.9$ & $79.4$ & $89.1$ & $2.0$ & $8.5$\\

         % coco_vse0_resnet_restval
         % --data_name coco --cnn_type resnet152 --use_restval --no_imgnorm
         \numrow{}\label{coco:VSEzRes} &
         \VSEzRes{} & \RCV{} &
$455.1$ &
$52.7$ & $83.0$ & $91.8$ & $1.0$ & $4.5$&
$36.0$ & $72.6$ & $85.5$ & $2.2$ & $7.7$
         \\
         {}\ref{coco:VSEppRes} &
         \VSEppRes{} & \RCV{} &
         $446.8$ &
         $58.3$ & $86.1$ & $93.3$ & $\mathbf{1.0}$ & $4.2$ &
         $43.6$ & $77.6$ & $87.8$ & $2.0$ & $7.8$
         \\

         % coco_vse0_resnet_restval
         % --data_name coco --cnn_type resnet152 --use_restval --no_imgnorm
         %  --finetune --num_epochs 15 --learning_rate .00002 --resume
        \numrow{}\label{coco:VSEzResFt} &
         \VSEzResFt{} & \RCV{} &
$470.6$ &
$56.0$ & $85.8$ & $93.5$ & $1.0$ & $4.0$&
$43.7$ & $79.4$ & $89.7$ & $2.0$ & $6.2$
         \\
          {}\ref{coco:VSEppResFt} &
         \VSEppResFt{} & \RCV{} &
         $\mathbf{478.6}$ &
         $\mathbf{64.6}$ & $\mathbf{90.0}$ & $\mathbf{95.7}$ & $\mathbf{1.0}$ 
         & $\mathbf{3.4}$ &
         $\mathbf{52.0}$ & $\mathbf{84.3}$ & $\mathbf{92.0}$ & $\mathbf{1.0}$ 
         & $\mathbf{6.1}$
         \\[-1mm]

        % instead of repeating the rows, we ensured Table 1,2 come together
         % {}\ref{coco:VSEppResFt} &
         % \VSEppResFt{} & \RCV{} &
         % $\mathbf{478.6}$ &
         % $\mathbf{64.6}$ & $\mathbf{90.0}$ & $\mathbf{95.7}$ & $\mathbf{1.0}$ 
         % & $\mathbf{3.4}$ &
         % $\mathbf{52.0}$ & $\mathbf{84.3}$ & $\mathbf{92.0}$ & $\mathbf{1.0}$ 
         % & $\mathbf{6.1}$
         % \\
    \end{tabular}
    }
    \vspace{.2cm}
    \caption{\small The effect of data augmentation and fine-tuning. We copy the 
    relevant results for \VSEpp{} from Table~\ref{tb:coco} to enable an easier 
    comparison.  Notice that after applying all the modifications, \VSEz{} 
    model reaches $56.0\%$ for $R@1$, while \VSEpp{} achieves $64.6\%$.}
    \label{tb:vse0}
    \vspace{-.4cm}
% \end{table*}

\end{table*}

The results on the \coco{} dataset are presented in Table~\ref{tb:coco}.  To 
understand the effect of training and algorithmic variations we report ablation 
studies for the baseline \VSEz{} (see Table~\ref{tb:vse0}).  Our best result 
with \VSEpp{} is achieved by using ResNet152 and fine-tuning the image encoder 
(row~\ref{coco:VSEppResFt}), where we see $21.2\%$ improvement in R@1 for 
caption retrieval and $21\%$ improvement in R@1 for image retrieval compared to 
\VSE{} (rows~\ref{coco:VSE} and~\ref{coco:VSEppResFt}).  Notice that using 
ResNet152 and fine-tuning can only lead to $12.6\%$ improvement using the 
\VSEz{} formulation (rows~\ref{coco:VSEzResFt} and~\ref{coco:VSE}), while our 
\MAX{} loss function brings a significant gain of $8.6\%$ 
(rows~\ref{coco:VSEppResFt} and~\ref{coco:VSEzResFt}).

%Although the improved loss can also improve the state-of-the-art models, our 
%results are already better than the best reported results. 

Comparing \VSEppResFt{} to the current state-of-the-art on \coco{}, {\em 
2WayNet}\/ (row~\ref{coco:VSEppResFt} and row~\ref{coco:twoway}), we see 
$8.8\%$ improvement in R@1 for caption retrieval and compared to 
{\em sm-LSTM}\/ (row~\ref{coco:VSEppResFt} and row~\ref{coco:smLSTM}), 
$11.3\%$ improvement in image retrieval.
We also report results on the full $5K$ test set of \coco{} in 
rows~\ref{coco:VSEppFt5} and~\ref{coco:VSEppResFt5}.

\emph{Effect of the training set}. We compare \VSEz{} and \VSEpp{} by 
incrementally improving the training data.  Comparing the models trained on 
\OC{} (rows~\ref{coco:VSE} and~\ref{coco:VSEppOC}), we only see $2.7\%$ 
improvement in R@1 for image retrieval but no improvement in caption retrieval 
performance. However, when we train using \RC{} (rows~\ref{coco:VSEppRC} 
and~\ref{coco:VSEzRC}) or \RCV{} (rows~\ref{coco:VSEppRCV} 
and~\ref{coco:VSEzRCV}), we see that \VSEpp{} gains an improvement of $5.9\%$ 
and $5.1\%$, respectively,  in R@1 for caption retrieval compared to \VSEz{}.  
This shows that \VSEpp{} can better exploit the additional data.

\emph{Effect of a better image encoding}. We also investigate the effect of 
a better image encoder on the models.  Row~\ref{coco:VSEppFt} and 
row~\ref{coco:VSEzFt} show the effect of fine-tuning the VGG19 image encoder. 
We see that the gap between \VSEz{} and \VSEpp{} increases to $6.1\%$. If we 
use ResNet152 instead of VGG19 (row~\ref{coco:VSEppRes} and 
row~\ref{coco:VSEzRes}), the gap is $5.6\%$. As for our best result, if we use 
ResNet152 and also fine-tune the image encoder (row~\ref{coco:VSEppResFt} and 
row~\ref{coco:VSEzResFt}) the gap becomes $8.6\%$. The increase in the 
performance gap shows that the improved loss of \VSEpp{} can better guide the 
optimization when a more powerful image encoder is used.

\subsection{Results on \fthk{}}

\begin{table*}[t!]
%\begin{table*}[t]
    \resizebox{\linewidth}{!}{
        \centering
    \begin{tabular}{c|c|c|HccccH|ccccH}
        \# &
        {\bf Model} & {\bf Trainset} & {\bf R Sum}
        & \multicolumn{5}{c|}{Caption Retrieval}
        & \multicolumn{5}{c}{Image Retrieval}\\
        & & & &
        {\bf R@1} & {\bf R@5} & {\bf R@10} & {\bf Med r} & {\bf Mean r} &
        {\bf R@1} & {\bf R@5} & {\bf R@10} & {\bf Med r} & {\bf Mean r}\\
        \hline
        \numrow{}\label{f30k:VSE} &
        \VSE{}~(\cite{kiros2014unifying}) & \OC{}  &
        $251.9$ &
        $23.0$ & $50.7$ & $62.9$ & $5$ & - &
        $16.8$ & $42.0$ & $56.5$ & $8$ & -
        \\
        \numrow{}\label{f30k:VSEgit} &
        \VSE{} (GitHub) & \OC{}  &
        $287.9$ &
        $29.8$ & $58.4$ & $70.5$ & $4$ & - &
        $22.0$ & $47.9$ & $59.3$ & $6$ & -
        \\
        \numrow{} &
        Embedding Net~(\cite{wang2017learning}) & \TC{} &
        - &
        $40.7$ & $69.7$ & $79.2$ & - &? &
        $29.2$ & $59.6$ & $71.7$ & - & -
        \\
        \numrow{} &
        DAN~(\cite{nam2016dual}) &? &
        - &
        $41.4$ & $73.5$ & $82.5$ & $2$ & - &
        $31.8$ & $61.7$ & $72.5$ & $3$ & -
        \\
        \numrow{} &
        sm-LSTM~(\cite{huang2016instance}) &? &
        $358.7$ &
        $42.5$ & $71.9$ & $81.5$ & $2$ & - &
        $30.2$ & $60.4$ & $72.3$ & $3$ & -
        \\
        \numrow{} &
        2WayNet~(\cite{eisenschtat2016linking}) & \TC{} &
        - &
        $49.8$ & $67.5$ & - & - & - &
        $36.0$ & $55.6$ & - & - & -
        \\
        \numrow{} &
        DAN (ResNet)~(\cite{nam2016dual}) &? &
        - &
        $\mathbf{55.0}$ & $\mathbf{81.8}$ & $\mathbf{89.0}$ & $\mathbf{1}$ 
        & - &
        $\mathbf{39.4}$ & $\mathbf{69.2}$ & $\mathbf{79.1}$ & $\mathbf{2}$ & -
        \\
        \hline
        % f30k_vse0
        % --data_name f30k_precomp --no_imgnorm
        \numrow{} &
        \VSEz{} & \OC{}  &
$294.3$ &
$29.8$ & $59.8$ & $71.9$ & $3.0$ & $24.7$ &
$23.0$ & $48.8$ & $61.0$ & $6.0$ & $35.3$
        \\

        % f30k_vse0_vggfull
        % --data_name f30k --cnn_type vgg19
        \numrow{}\label{f30k:VSEzRC}&
        \VSEz{} & \RC &
$298.7$ &
%$54.2$ &
$31.6$ & $59.3$ & $71.7$ & $4.0$ & $25.3$&
%$45.4$ &
$21.6$ & $50.7$ & $63.8$ & $5.0$ & $30.0$
        \\

        % f30k_vse++
        % --data_name f30k_precomp --max_violation
        \numrow{} &
        \VSEpp{} & \OC{}  &
$291.3$ &
$31.9$ & $58.4$ & $68.0$ & $4.0$ & $30.9$ &
$23.1$ & $49.2$ & $60.7$ & $6.0$ & $35.6$
        \\

       % f30k_vse++_vggfull
        % --data_name f30k --cnn_type vgg19 --max_violation
        \numrow{}\label{f30k:VSEppRC}&
        \VSEpp{} & \RC &
        $326.3$ &
        $38.6$ & $64.6$ & $74.6$ & $2.0$ & $21.3$ &
        $26.8$ & $54.9$ & $66.8$ & $4.0$ & $30.0$\\

        % f30k_vse0_vggfull_finetune
        % --data_name f30k --cnn_type vgg19 --no_imgnorm --finetune
        %  --num_epochs 15 --learning_rate 0.00002 --resume
\numrow{}&
\VSEzFt{} & \RC &
$333.9$ & 
$37.4$ & $65.4$ & $77.2$ & $3.0$ & $16.2$&
$26.8$ & $57.6$ & $69.5$ & $4.0$ & $22.9$
\\

        % f30k_vse++_vggfull_finetune
        % --data_name f30k --cnn_type vgg19 --max_violation --finetune 
        %  --num_epochs 15 --learning_rate .00002 --resume
        \numrow{} &
        \VSEppFt{} & \RC &
$350.9$ &
$41.3$ & $69.1$ & $77.9$ & $2.0$ & $18.2$ &
$31.4$ & $60.0$ & $71.2$ & $3.0$ & $24.9$
        \\

        % f30k_vse0_resnet
        % --data_name f30k --cnn_type resnet152 --no_imgnorm
\numrow{}&
\VSEzRes{} & \RC &
$324.2$ & 
$36.6$ & $67.3$ & $78.4$ & $3.0$ & $12.6$&
$23.3$ & $52.6$ & $66.0$ & $5.0$ & $25.4$
\\
        % f30k_vse++_resnet
        % --data_name f30k --cnn_type resnet152 --max_violation
        \numrow{} &
        \VSEppRes{} & \RC &
$363.1$ &
% $65.9$ &
$43.7$ & $71.9$ & $82.1$ & $2.0$ & $12.5$&
% $55.1$ &
$32.3$ & $60.9$ & $72.1$ & $3.0$ & $22.2$
        \\

        % f30k_vse0_resnet_finetune
        % --data_name f30k --cnn_type resnet152 --no_imgnorm --finetune
        %  --num_epochs 15 --learning_rate .00002 --resume
\numrow{}&
\VSEzResFt{} & \RC &
$367.7$ & 
$42.1$ & $73.2$ & $84.0$ & $2.0$ & $11.1$&
$31.8$ & $62.6$ & $74.1$ & $3.0$ & $18.6$
\\
        % f30k_vse++_resnet_finetune
        % --data_name f30k --cnn_type resnet152 --max_violation --finetune 
        %  --num_epochs 15 --learning_rate .00002 --resume
        \numrow{}\label{f30k:VSEppResFt} &
        \VSEppResFt{} & \RC &
$409.7$ &
% $73.5$ &
$52.9$ & $80.5$ & $87.2$ & $1.0$ & $9.5$&
% $63.0$ &
$\mathbf{39.6}$ & $\mathbf{70.1}$ & $\mathbf{79.5}$ & $\mathbf{2.0}$ 
& $\mathbf{16.6}$
        \\[-1mm]

   \end{tabular}
    }
    \vspace{.2cm}
    \caption{Results on the \fthk{} dataset.}
    \label{tb:f30k}
    \vspace{-.4cm}
%\end{table*}

\end{table*}

Tables~\ref{tb:f30k} summarizes the performance on \fthk{}.  We obtain $23.1\%$ 
improvement in R@$1$ for caption retrieval and $17.6\%$ improvement in R@$1$ 
for image retrieval (rows~\ref{f30k:VSE} and~\ref{f30k:VSEppResFt}). We 
observed that \VSEpp{} over-fits when trained with the pre-computed features of 
\OC{}. The reason is potentially the limited size of the \fthk{} training set.  
As explained in Sec.~\ref{sec:train_detail}, we select a snapshot of the model 
before over-fitting occurs, based on performance with the validation set.  
Over-fitting does not occur when the model is trained using the \RC{} training 
data.  Our results show the improvements incurred by our \MAX{} loss persist 
across datasets, as well as across models.

\subsection{Improving Order Embeddings}
\label{sec:order}

% \david{We have never discussed order embeddings so this needs more 
% background. Do it however you'd like but remember that the reader is not
% expecting this twist.}

Given the simplicity of our approach, our proposed loss function can complement 
the recent approaches that use more sophisticated model architectures or 
similarity functions. Here we demonstrate the benefits of the \MAX{} loss by 
applying it to another approach to joint embeddings called 
order-embeddings~\cite{vendrov2015order}.  The main difference with the 
formulation above is the use of an asymmetric similarity function, i.e., 
$s(i,c)=-\|\max(0, g(c; W_g, \theta_\psi) - f(i; W_f, \theta_\phi)) \|^2$.  
Again, we simply replace their use of the \SUM{} loss by our \MAX{} loss.

Like their experimental setting, we use the training set \TCV{}.  For our 
\orderpp{}, we use the same learning schedule and margin as our other 
experiments.  However, we use their training settings to train \orderz{}. 
We start training with a learning rate of $0.001$ for $15$ epochs 
and lower the learning rate to $0.0001$ for another $15$ epochs. 
Like \cite{vendrov2015order} we use a margin of $0.05$.  
Additionally, \cite{vendrov2015order} takes the absolute value of 
embeddings before computing the similarity function which we replicate 
only for \orderz{}.

Table~\ref{tb:order} reports the results when the \SUM{} loss 
is replaced by the \MAX{} loss.  We replicate their results using 
our \orderz{} formulation and get slightly better results 
(row~\ref{order:order} and row~\ref{order:orderz}). We observe $4.5\%$ 
improvement from \orderz{} to \orderpp{} in R@$1$ for caption retrieval 
(row~\ref{order:orderz} and row~\ref{order:orderpp}).  Compared to the 
improvement from \VSEz{} to \VSEpp{}, where the improvement on the \TCV{} 
training set is $1.8\%$, we gain an even higher improvement here.
This shows  that the \MAX{} loss can potentially improve numerous similar 
loss functions used in retrieval and ranking tasks.

\begin{table}[h]
    % \begin{tabular}{>{\centering\arraybackslash}m{0.65\linewidth}
    %     >{\centering\arraybackslash}m{0.3\linewidth}
    %     }
    %     \begin{tabular}{>{\centering\arraybackslash}m{0.65\linewidth}
    %         >{\centering\arraybackslash}m{0.3\linewidth}
    %         }
    %         \begin{tabular}{lcc}
    %\begin{table*}[t]
   \resizebox{\linewidth}{!}{\centering
  %\small \addtolength{\tabcolsep}{1.pt}
     \begin{tabular}{c|c|HHccccH|ccccH}
         \# &
         {\bf Model} & {\bf Trainset} & {\bf R Sum}
         & \multicolumn{5}{c|}{Caption Retrieval}
         & \multicolumn{5}{c}{Image Retrieval}\\
         & & & &
         {\bf R@1} & {\bf R@5} & {\bf R@10} & {\bf Med r} & {\bf Mean r} &
         {\bf R@1} & {\bf R@5} & {\bf R@10} & {\bf Med r} & {\bf Mean r}\\
         \hline
         & & \multicolumn{12}{c}{{\cellcolor[gray]{0.8}\bf 1K Test Images}}\\
         \hline

         \numrow{}\label{order:order} &
         \order{}~(\cite{vendrov2015order})& \TCV{} &
         - &
         $46.7$ & - & $88.9$ & $2.0$ & $5.7$ &
         $37.9$ & - & $85.9$ & $2.0$ & $8.1$\\
         \hline

%         % coco_order0
%         % --data_name coco_precomp --metric order --use_abs --margin .05 
%         %  --learning_rate .001
%         \orderz{} & \OC{} (1 fold) &
%         $379.4$ &
%         $39.8$ & $72.2$ & $83.8$ & $2.0$ & $9.2$ &
%         $33.3$ & $68.7$ & $81.7$ & $3.0$ & $10.3$\\
%
%         % coco_order++
%         % --data_name coco_precomp --metric order --max_violation 
%         \orderpp{} & \OC{} (1 fold) &
%         $404.2$ &
%         $46.3$ & $77.6$ & $87.2$ & $2.0$ & $7.2$ &
%         $37.3$ & $72.0$ & $83.8$ & $2.0$ & $10.1$ \\
%
%         \hline
%
         % coco_vse0_10crop
         % --data_name 10crop --no_imgnorm
         \numrow{}\label{order:vsez} &
         \VSEz{} & \TCV{} &
         $417.0$ &
         $49.5$ & $81.0$ & $90.0$ & $1.8$ & $5.1$ &
         $38.1$ & $73.3$ & $85.1$ & $2.0$ & $8.4$\\

         % coco_order0_10crop
         % --data_name 10crop --metric order --use_abs --learning_rate .001 
         % --margin .05
         \numrow{}\label{order:orderz} &
         \orderz{} & \TCV{} &
         $421.3$ &
         $48.5$ & $80.9$ & $90.3$ & $1.8$ & $5.2$ &
         $39.6$ & $75.3$ & $86.7$ & $2.0$ & $7.4$
         \\

         % coco_vse++_10crop
         % --data_name 10crop --max_violation
         \numrow{}\label{order:vsepp}&
         \VSEpp{} & \TCV{} &
         $425.9$ &
         $51.3$ & $82.2$ & $91.0$ & $1.2$ & $5.0$ &
         $40.1$ & $75.3$ & $86.1$ & $2.0$ & $10.5$
         \\

         % coco_order++_10crop
         % --data_name 10crop --metric order --max_violation
        \numrow{}\label{order:orderpp}&
         \orderpp{} & \TCV{} &
         $436.0$ &
         ${\bf 53.0}$ & $83.4$ & ${\bf 91.9}$ & ${\bf 1.0}$ & $4.6$ &
         ${\bf 42.3}$ & $77.4$ & ${\bf 88.1}$ & $2.0$ & $8.2$
         \\[-1mm]
    \end{tabular}
    }
%\end{table*}

    \vspace{.2cm}
        \captionof{table}{Comparison on \coco{}. Training set for all the rows is 
        \TCV{}.}
        \label{tb:order}
    %     \end{tabular}
    %     \end{tabular}
    %     \end{tabular}
    \vspace{-.4cm}
\end{table}

\subsection{Behavior of Loss Functions}

\begin{figure}[h]
    \centering
    %\begin{subfigure}{.47\textwidth}
        \includegraphics[width=.5\linewidth]{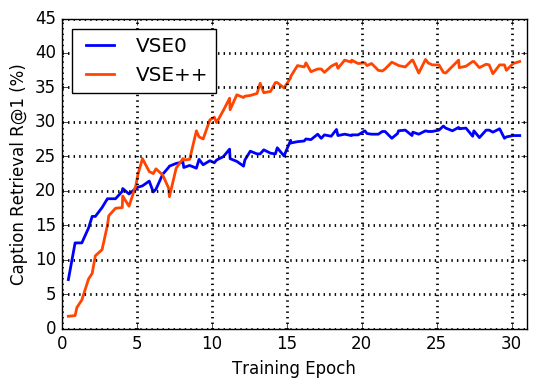}
        %\caption{}\label{fig:sum_vs_max}
    %\end{subfigure}
        \caption{Analysis of the behavior of the \MAX{} loss on the \fthk{} 
        dataset training with \RC{}. This figure compares the \SUM{} loss to 
        the \MAX{} loss (Table~\ref{tb:f30k}, row~\ref{f30k:VSEzRC} and 
        row~\ref{f30k:VSEppRC}). Notice that, in the first $5$ epochs the 
        \SUM{} loss achieves a better performance, however, from there-on the 
        \MAX{} loss leads to much higher recall rates.}
        \label{fig:sum_vs_max}
        %\label{fig:svm_neg}
\end{figure}

We  observe that the \MAX{} loss can take a few epochs to `warm-up' 
during training.  Fig.~\ref{fig:sum_vs_max} depicts such behavior on the 
\fthk{} dataset using \RC{}.  Notice that the \SUM{} loss starts off 
faster, but after approximately $5$ epochs \MAX{} loss surpasses \SUM{} loss.  
To explain this, the \MAX{} loss depends on a smaller set of triplets 
compared to the \SUM{} loss.  Early in training the gradient of the \MAX{} 
loss is  influenced by a relatively small set of triples.  As such, it can 
take more iterations to train a model with the \MAX{} loss. We explored a simple 
form of curriculum learning (\cite{bengio2009curriculum}) to speed-up the  
training. We start training with the \SUM{} loss for a few epochs, then 
switch to the \MAX{} loss for the rest of the training. However, it did 
not perform much better than training solely with the \MAX{} loss.

%We also investigated how the \MAX{} loss affects the distribution of distances 
%for hard negatives. Fig.~\ref{fig:dists_f30k} shows that for the \MAX{} loss, 
%the distribution of distances is skewed to the left in the middle of the 
%training while for the \SUM{} loss the distribution is approximately 
%symmetric.  This is an expected behavior because the \MAX{} loss 
%disproportionately pushes the hardest negatives away. A model trained with the 
%\MAX{} loss at epoch $10$, has fewer pairs with at least a cosine similarity 
%of $0.4$ compared to another model trained with the \SUM{} loss. At Epoch 
%$30$, this margin is reduced to $0.3$.
%
%\begin{figure}[h]
%    \includegraphics[width=\linewidth]{images/dists_f30k.png}
%        \caption{Distribution of distances for \fthk{} training set averaged 
%        over $10$ fixed examples over the course of training. The distribution 
%        for the \MAX{} loss and the \SUM{} loss is shown for snapshots of the 
%        model at epoch $10$ and $30$. For comparison, the distribution for 
%        a randomly initialized model is also depicted.}
%        \label{fig:dists_f30k}
%\end{figure}

In \cite{schroff2015facenet}, it is reported that with a mini-batch size of 
$1800$,  training  is extremely slow. We experienced similar behavior 
with large mini-batches up to $512$. However, mini-batches of 
size $128$ or $256$ exceeded the performance of the \SUM{} loss within 
the same 
training time.

% Based on this observation, in Fig.~\ref{fig:nonzero}, we explored a simple 
% form of curriculum learning (\cite{bengio2009curriculum}) to speed-up the 
% training.  We start training with the \SUM{} loss for a few epochs, then 
% switch to the \MAX{} loss for the rest of the training.  
% Fig.~\ref{fig:nonzero} shows that, for large mini-batch sizes, the number of 
% non-zero terms in the loss does not go down for many epochs.  The model 
% quickly jumps to a state similar to a model that was trained from scratch 
% with the \MAX{} loss.  Intuitively, this can happen because fixing hard 
% negatives might need significant changes to the model rather than incremental 
% improvements.

% \input{neg_size}

\subsection{Examples of Hard Negatives}

\begin{figure*}[t!]
\centering
\scriptsize
\begin{tabular}[h]{>{\centering\arraybackslash}m{0.22\linewidth}
>{\centering\arraybackslash}m{0.22\linewidth}
>{\centering\arraybackslash}m{0.22\linewidth}
>{\centering\arraybackslash}m{0.22\linewidth}}
\makecell[{{p{\linewidth}}}]{\includegraphics[width=\linewidth, height=\linewidth]{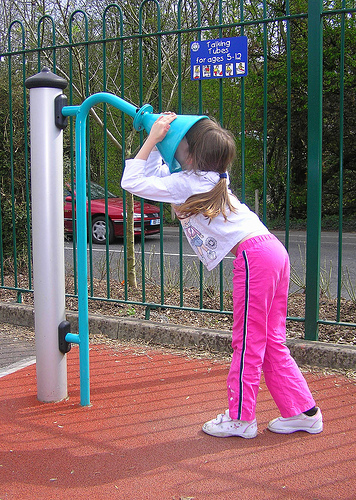}\\[1mm]
{\bf GT}: A little girl wearing pink pants, pink and white tennis shoes and a white shirt with a little girl on it puts her face in a blue Talking Tube. \\[1mm]
{\bf HN}: [0.26] Blond boy jumping onto deck. \\[9mm]
}
&
\makecell[{{p{\linewidth}}}]{\includegraphics[width=\linewidth, height=\linewidth]{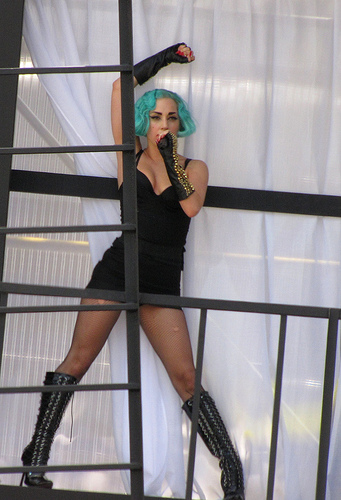}\\[1mm]
{\bf GT}: A teal-haired woman in a very short black dress, pantyhose, and boots standing with right arm raised and left hand obstructing her mouth in microphone-singing fashion is standing. \\[1mm]
{\bf HN}: [0.08] Two dancers in azure appear to be performing in an alleyway. \\[1mm]
}
&
\makecell[{{p{\linewidth}}}]{\includegraphics[width=\linewidth, height=\linewidth]{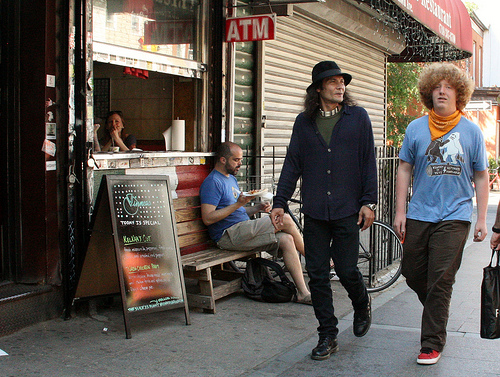}\\[1mm]
{\bf GT}: Two men, one in a dark blue button-down and the other in a light blue tee, are chatting as they walk by a small restaurant. \\[1mm]
{\bf HN}: [0.41] Two men with guitars strapped to their back stand on the street corner with two other people behind them. \\[1mm]
}
&
\makecell[{{p{\linewidth}}}]{\includegraphics[width=\linewidth, height=\linewidth]{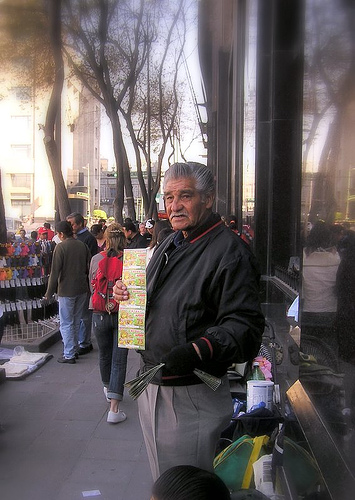}\\[1mm]
{\bf GT}: A man wearing a black jacket and gray slacks, stands on the sidewalk holding a sheet with something printed on it in his hand. \\[1mm]
{\bf HN}: [0.26] Two men with guitars strapped to their back stand on the street corner with two other people behind them. \\[1mm]
}
\\
\makecell[{{p{\linewidth}}}]{\includegraphics[width=\linewidth, height=\linewidth]{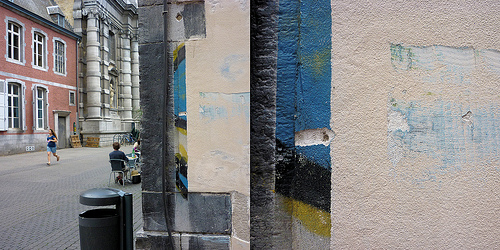}\\[1mm]
{\bf GT}: There is a wall of a building with several different colors painted on it and in the distance one person sitting down and another walking. \\[1mm]
{\bf HN}: [0.06] A woman with luggage walks along a street in front of a large advertisement. \\
}
&
\makecell[{{p{\linewidth}}}]{\includegraphics[width=\linewidth, height=\linewidth]{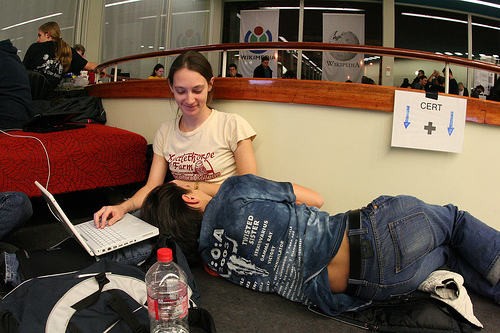}\\[1mm]
{\bf GT}: A man is laying on a girl's lap, she is looking at him, she also has her hand on her notebook computer. \\[1mm]
{\bf HN}: [0.18] A woman sits on a carpeted floor with a baby. \\[11mm]
}
&
\makecell[{{p{\linewidth}}}]{\includegraphics[width=\linewidth, height=\linewidth]{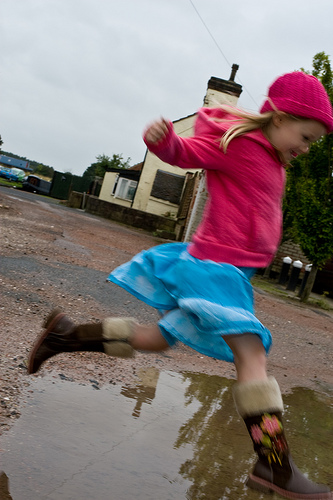}\\[1mm]
{\bf GT}: A young blond girl in a pink sweater, blue skirt, and brown boots is jumping over a puddle on a cloudy day. \\[1mm]
{\bf HN}: [0.51] An Indian woman is sitting on the ground, amongst drawings, rocks and shrubbery. \\[5mm]
}
&
\makecell[{{p{\linewidth}}}]{\includegraphics[width=\linewidth, height=\linewidth]{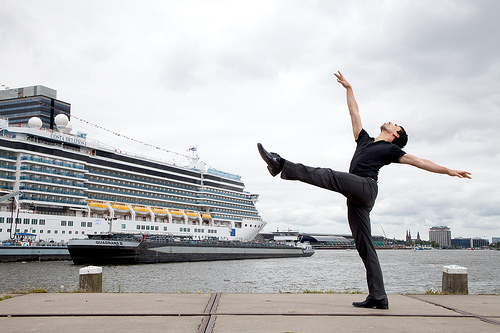}\\[1mm]
{\bf GT}: One man dressed in black is stretching his leg up in the air, behind him is a massive cruise ship in the water. \\[1mm]
{\bf HN}: [0.24] A topless man straps surfboards on top of his blue car. \\[5mm]
}
\\
\end{tabular}
    \vspace{.3cm}
\caption{Examples from the \fthk{} training set along with their hard negatives 
    in a random mini-batch according to the loss of a trained \VSEpp{} model.  
    The value in brackets is the cost of the hard negative and is in the range 
    $[0, 2]$ in our implementation.  HN is the hardest negative in a random 
    sample of size $128$.  GT is the positive caption used to compute the cost 
    of NG.}\label{fig:sample_hard_negatives}
\end{figure*}

Fig.~\ref{fig:sample_hard_negatives} shows the hard negatives in a random 
mini-batch. These examples illustrate that hard negatives from a mini-batch can 
provide useful gradient information.

\section{Conclusion}
This paper focused on learning visual-semantic embeddings for
cross-modal, image-caption retrieval.  Inspired by structured prediction, 
we proposed a new loss based on violations incurred by relatively hard 
negatives compared to current methods that used expected errors
(\cite{kiros2014unifying,vendrov2015order}). We performed experiments  
on the \coco{} and \fthk{} datasets and showed that our proposed loss 
significntly  improves performance on these datasets. We observed that 
the improved loss can better guide a more powerful image encoder, 
ResNet152, and also guide better when fine-tuning an image encoder. 
With all modifications, our \VSEpp{} model achieves state-of-the-art 
performance on the \coco{} dataset, and is slightly below the best 
recent model on the \fthk{} dataset.  Our proposed loss function can 
be used to train more sophisticated models that have been using a 
similar ranking loss for training.

\section*{Acknowledgements}
This work was supported in part by funding to DF from NSERC Canada, 
the Vector Institute, and the Learning in Brains and Machines Program 
of the Canadian Institute for Advanced Research.

%{\small
%\bibliographystyle{ieee}
\bibliography{bib}
%}

% For arxiv copy
\ifdefined\arxivcopy
\newpage
\appendix
\part*{Appendix}
\setcounter{figure}{0}
\renewcommand\thefigure{\thesection.\arabic{figure}}
\section{Examples of Hard Negatives}

Fig.~\ref{fig:sample_outputs} compares the outputs of \VSEpp{} and \VSEz{} for 
a few examples.

\begin{figure*}[h!]
\centering
\scriptsize
\begin{tabular}[h]{>{\centering\arraybackslash}m{0.22\linewidth}
>{\centering\arraybackslash}m{0.22\linewidth}
>{\centering\arraybackslash}m{0.22\linewidth}
>{\centering\arraybackslash}m{0.22\linewidth}}
\makecell[{{p{\linewidth}}}]{\includegraphics[width=\linewidth, height=\linewidth]{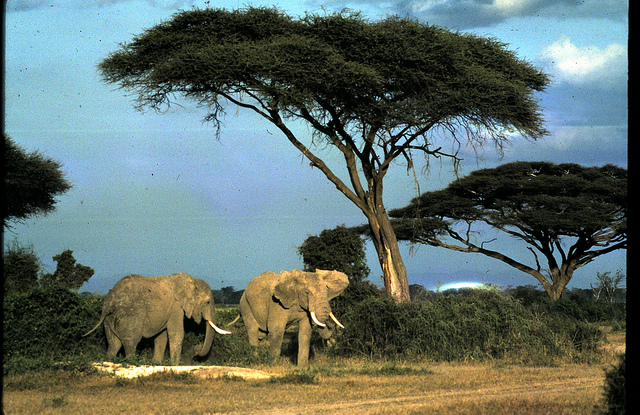}\\[1mm]
{\bf GT}: Two elephants are standing by the trees in the wild.  \\[1mm]
{\bf \VSEz{}}: [9] Three elephants kick up dust as they walk through the flat by the bushes. \\[1mm]
{\bf \VSEpp{}}: [1] A couple elephants walking by a tree after sunset.\\[2mm]
}
&
\makecell[{{p{\linewidth}}}]{\includegraphics[width=\linewidth, height=\linewidth]{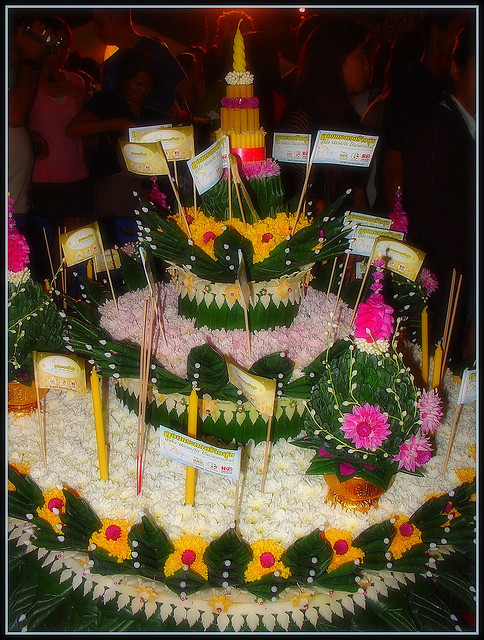}\\[1mm]
{\bf GT}: A large multi layered cake with candles sticking out of it.  \\[1mm]
{\bf \VSEz{}}: [1] A party decoration containing flowers, flags, and candles. \\[1mm]
{\bf \VSEpp{}}: [1] A party decoration containing flowers, flags, and candles. \\[4mm]
}
&
\makecell[{{p{\linewidth}}}]{\includegraphics[width=\linewidth, height=\linewidth]{images/COCO_val2014_000000408449.jpg}\\[1mm]
{\bf GT}: The man is walking down the street with no shirt on. \\[1mm]
{\bf \VSEz{}}: [24] A person standing on a skate board in an alley. \\[1mm]
{\bf \VSEpp{}}: [10] Two young men are skateboarding on the street. \\[4mm]
}
&
\makecell[{{p{\linewidth}}}]{\includegraphics[width=\linewidth, height=\linewidth]{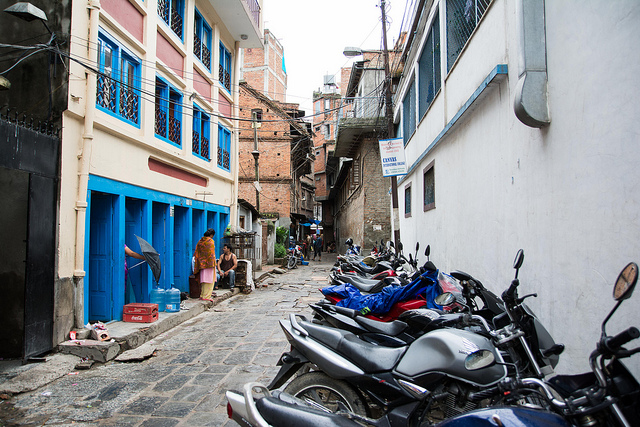}\\[1mm]
{\bf GT}: A row of motorcycles parked in front of a building. \\[1mm]
{\bf \VSEz{}}: [2] a parking area for motorcycles and bicycles along a street \\[1mm]
{\bf \VSEpp{}}: [1] A number of motorbikes parked on an alley \\[4mm]
}
\\
\makecell[{{p{\linewidth}}}]{\includegraphics[width=\linewidth, height=\linewidth]{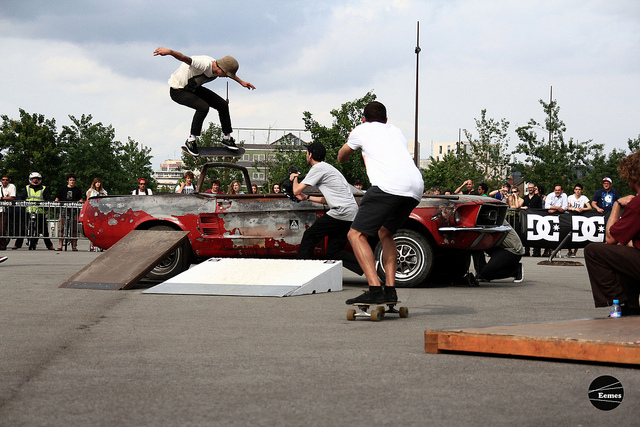}\\[1mm]
{\bf GT}: some skateboarders doing tricks and people watching them \\[1mm]
{\bf \VSEz{}}: [39] Young skateboarder displaying skills on sidewalk near field. \\[1mm]
{\bf \VSEpp{}}: [3] Two young men are outside skateboarding together. \\
}
&
\makecell[{{p{\linewidth}}}]{\includegraphics[width=\linewidth, height=\linewidth]{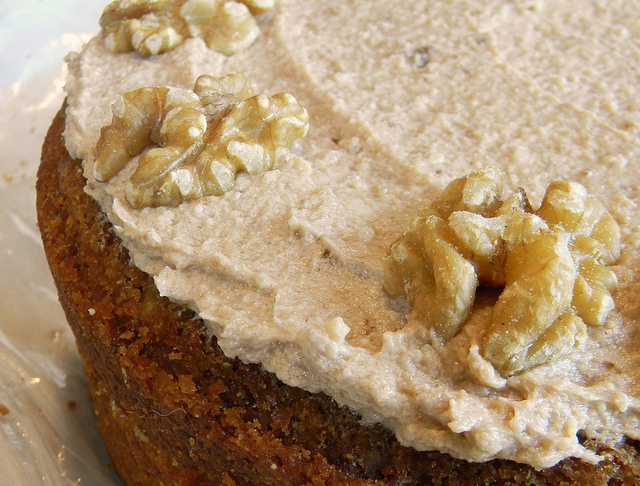}\\[1mm]
{\bf GT}: a brown cake with white icing and some walnut toppings \\[1mm]
{\bf \VSEz{}}: [6] A large slice of angel food cake sitting on top of a plate. \\[1mm]
{\bf \VSEpp{}}: [16] A baked loaf of bread is shown still in the pan. \\
}
&
\makecell[{{p{\linewidth}}}]{\includegraphics[width=\linewidth, height=\linewidth]{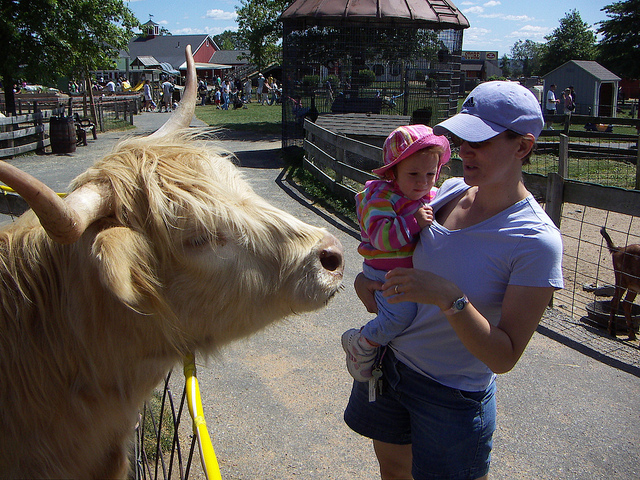}\\[1mm]
{\bf GT}: A woman holding a child and standing near a bull. \\[1mm]
{\bf \VSEz{}}: [1] A woman holding a child and standing near a bull. \\[1mm]
{\bf \VSEpp{}}: [1] A woman holding a child looking at a cow. \\
}
&
\makecell[{{p{\linewidth}}}]{\includegraphics[width=\linewidth, height=\linewidth]{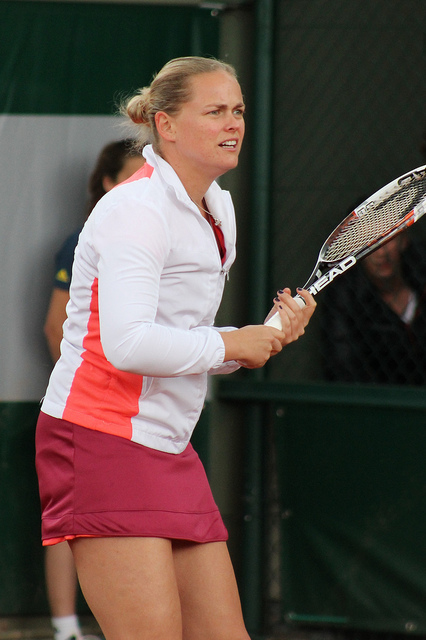}\\[1mm]
{\bf GT}: A woman in a short pink skirt holding a tennis racquet. \\[1mm]
{\bf \VSEz{}}: [6] A man playing tennis and holding back his racket to hit the ball. \\[1mm]
{\bf \VSEpp{}}: [1] A woman is standing while holding a tennis racket. \\
}
\end{tabular}
    \vspace{.3cm}
\caption{Examples of \coco{} test images and the top 1 retrieved captions for \VSEz{} and \VSEpp{} 
(ResNet)-finetune. The value in brackets is the rank of the highest ranked 
ground-truth caption. GT is a sample from the ground-truth captions.}
    \label{fig:sample_outputs}
\end{figure*}

\fi

\end{document}